%% file: main.tex
\documentclass[10pt,twocolumn,letterpaper]{article}

\usepackage[pagenumbers]{cvpr} 

\input{preamble}

\definecolor{cvprblue}{rgb}{0.21,0.49,0.74}
\usepackage[pagebackref,breaklinks,colorlinks,allcolors=cvprblue]{hyperref}

\def\paperID{4626} 
\def\confName{CVPR}
\def\confYear{2026}

\title{SuperClip Pyramid for Video Temporal Grounding}

\author{
Sanghyeok Lee\textsuperscript{\rm 1,2}\thanks{Work done at KAIST}\hspace{0.2cm}
Juyeon Ko\textsuperscript{\rm 3}\hspace{0.2cm}
Joonmyung Choi\textsuperscript{\rm 4}\hspace{0.2cm}
Jongha Kim\textsuperscript{\rm 4}\\
Dohwan Ko\textsuperscript{\rm 4}\hspace{0.2cm}
Ji Soo Lee\textsuperscript{\rm 1}\hspace{0.2cm}
Hyunwoo J. Kim\textsuperscript{\rm 1}\thanks{Corresponding author.\vspace{-0.5cm}}\vspace{0.2cm} \\
\textsuperscript{\rm 1}KAIST\hspace{0.8cm}
\textsuperscript{\rm 2}TwelveLabs\hspace{0.8cm} 
\textsuperscript{\rm 3}KT\hspace{0.8cm}
\textsuperscript{\rm 4}Korea Univertsity\vspace{0.3cm}
}

\begin{document}
\maketitle

\input{section/abstract}
\input{section/intro}
\input{section/relwork}
\input{section/method}

\input{section/exp}
\input{section/conclusion}

\appendix
\input{section/appendix}
{
    \small
    \bibliographystyle{ieeenat_fullname}
    \bibliography{main}
}


\end{document}

%% file: preamble.tex
\usepackage{url}            
\usepackage{booktabs}       
\usepackage{amsfonts}       
\usepackage{nicefrac}       
\usepackage{microtype}      
\usepackage{xcolor}         

\definecolor{LightGray}{gray}{0.9}

\usepackage[table]{xcolor}  
\usepackage{amsmath}
\usepackage{multirow}
\usepackage{adjustbox}
\usepackage{bbm}     
\usepackage{subcaption}



%% file: section/abstract.tex
\begin{abstract}
With the rapid expansion of online video content, video temporal grounding has become essential for localizing moments based on user-specific queries, comprising two subtasks: moment retrieval (MR) and highlight detection (HD).
While recent approaches adopt DETR-style decoders or temporal pyramid architectures, they rely on single-scale feature maps or uniformly downsampled representations, limiting their ability to capture semantic boundaries and variable-length events.
In this paper, we present \textbf{SuperClip Video Temporal Grounding (SCVTG)} built on a multi-scale temporal hierarchy that better captures distinctive events at various granularities.
SCVTG constructs a SuperClip pyramid by grouping adjacent clips into semantically coherent segments, termed SuperClips.
Further, to aggregate these SuperClips having variable duration and temporal location, we introduce a temporally adaptive kernel that aggregates SuperClips through temporally conditioned weights, enabling flexible aggregation.
On top of this, we design relative saliency losses to guide the model in learning inter-clip importance, enhancing its ability to distinguish relevant events.
Extensive experiments show that SCVTG achieves the best performance across five video temporal grounding benchmarks with promising gains.
\end{abstract}

%% file: section/intro.tex
\input{figures/graph}
\input{figures/intro}
\section{Introduction}
With the unprecedented growth of online video content, there has been a growing interest in accurately identifying and localizing the relevant moments based on user-specific natural-language queries.
To this end, video temporal grounding has emerged as a task aimed at localizing video clips corresponding to user queries, enabling models to extract semantically relevant segments from untrimmed videos.
Video temporal grounding is typically decomposed into two closely related subtasks: moment retrieval (MR) and highlight detection (HD).
Moment retrieval~\cite{lei2021detecting, GaoSYN17, RegneriRWTSP13} explicitly focuses on localizing all temporal boundaries, \textit{e.g.}, the start and end timestamps of video segments, that match the query.
In parallel, highlight detection~\cite{lei2021detecting, song2015tvsum, sun2014ranking} assigns saliency scores to each clip to indicate their relevance.

Prior works~\cite{lei2021detecting,moon2023query,sun2024tr, moon2023correlation} jointly solve these two tasks, leveraging cross-modal information exchange between video and text, followed by DETR~\cite{carion2020end}-style decoders that predict the temporal window given video-text pairs.
Recent studies~\cite{liu2024r,cao2025flashvtg} have leveraged a temporal pyramid to capture hierarchical temporal semantics.
Similar to FPNs~\cite{ghiasi2019fpn,lin2017feature,tan2020efficientdet} in object detection, these temporal pyramids represent video content at multi-scale resolutions, thereby improving the localization accuracy at more diverse temporal granularities, from fine-grained actions to long-duration events with extended contexts. 
However, most existing methods either depend on single-scale representations or uniformly downsampled video clips at fixed intervals using convolutional networks, without considering the semantic boundaries. 
Intuitively, events in real-world videos vary in duration and rarely conform to fixed temporal lengths.
As a result, they are prone to merging distinct events into a single segment or splitting continuous events into different segments.
Further, even if these video clips can be successfully grouped into coherent event units, downsampling these groups with varying durations and group sizes remains non-trivial, \eg, defining dedicated kernels for all possible combinations of temporal durations and positions is computationally prohibitive, while simple pooling, being permutation-invariant, loses the ordering of the constituent clips.

In this paper, we introduce \textbf{S}uper\textbf{C}lip \textbf{V}ideo \textbf{T}emporal \textbf{G}rounding (\textbf{SCVTG}), a unified framework that tackles moment retrieval and highlight detection, effectively capturing distinctive events with dynamic temporal granularities.
Inspired by \emph{Superpixels}~\cite{ren2003learning}, SCVTG hierarchically groups adjacent video clips into semantically coherent segments, termed \emph{SuperClips}, forming a SuperClip pyramid.
As illustrated in Figure~\ref{fig:intro}, these SuperClips act as a region proposal that likely corresponds to the same underlying event.
Further, to aggregate variable-length SuperClips, we introduce a concept of temporally adaptive kernel, where each constituent clip is projected using its dedicated kernel weights conditioned on the corresponding temporal information.
This formulation not only enables flexible aggregation over segments of varying lengths but also efficiently incorporates temporal context into the aggregation process.
Finally, we introduce relative saliency loss, encouraging the model to better reflect the relevance of the clips within the entire video.
Extensive experiments demonstrate that SCVTG achieves a significant improvement in five benchmarks on video temporal grounding tasks, including both moment retrieval and highlight detection: QVHighlights~\cite{lei2021detecting}, Charades-STA~\cite{GaoSYN17}, TACoS~\cite{RegneriRWTSP13}, TVSum~\cite{song2015tvsum}, and YouTube-HL~\cite{sun2014ranking}. 
In summary, our contributions are threefold:
 \begin{itemize}
     \item We propose a novel architecture for video temporal grounding that leverages SuperClips, which group adjacent clips into semantically coherent segments for more precise event localization.
     \item We introduce a temporally adaptive kernel reflecting the temporal information of constituent clips, enabling flexible aggregation of variable-length SuperClips.
     \item By training SCVTG with the proposed relative-saliency losses, we deliver the new state-of-the-art performances across five video temporal grounding benchmarks.
 \end{itemize}

%% file: figures/graph.tex
\begin{figure}[t!]
\centering
\includegraphics[width=0.9\linewidth]{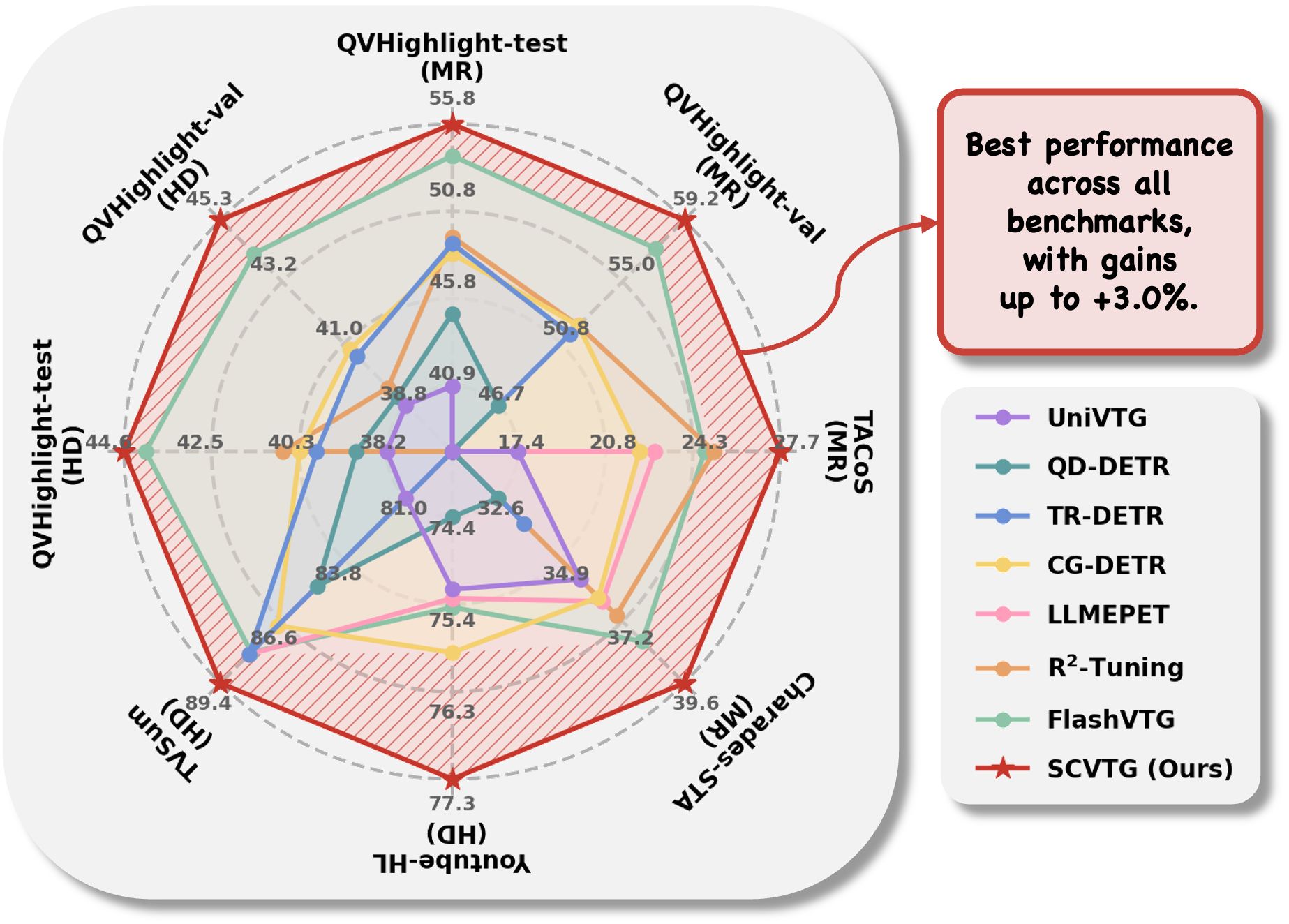}
\caption{
\textbf{Performance comparison.} SCVTG outperforms all baseline models, demonstrating its effectiveness in video temporal grounding.
For moment retrieval (MR) and highlight detection (HD), we use R1@0.7 and mAP@Avg, respectively. 
}
\label{fig:graph}
\vspace{-15pt}
\end{figure}

%% file: figures/intro.tex
\begin{figure*}[t!]
\centering
\includegraphics[width=0.9\textwidth]{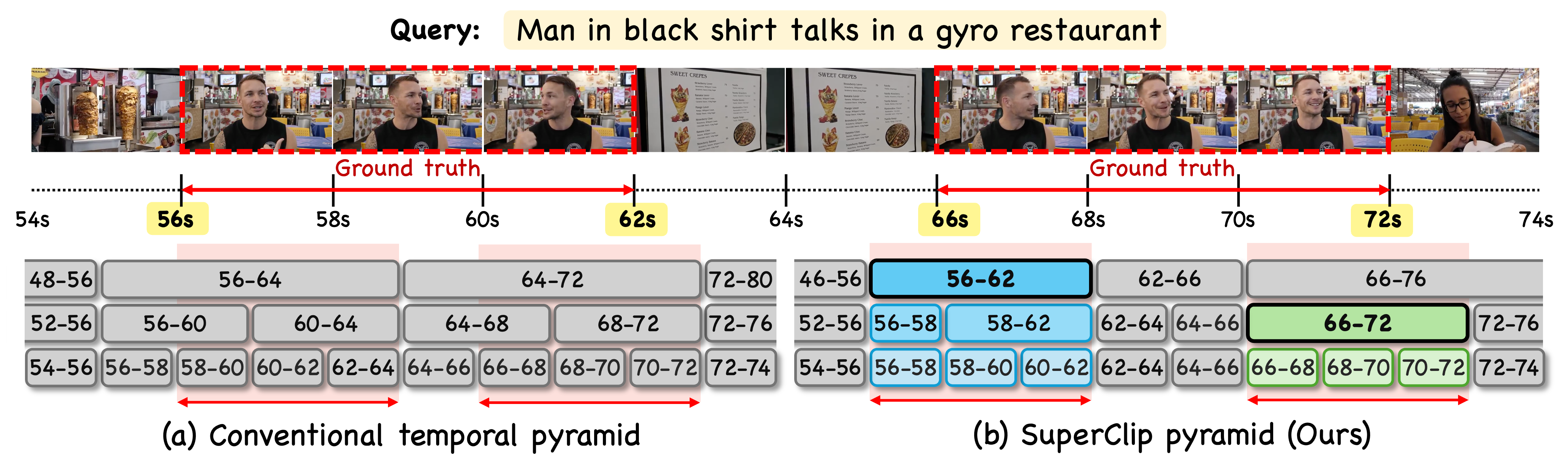}
\vspace{-10pt}

\caption{
\textbf{Comparison between (a) conventional temporal pyramid and (b) SuperClip pyramid.}
Unlike conventional temporal pyramids that segment videos at uniform intervals, our SuperClip pyramid adaptively groups adjacent clips into semantically coherent segments.
}
\label{fig:intro}
\vspace{-15pt}
\end{figure*}

%% file: section/relwork.tex
\section{Related Works}
\textbf{Video temporal grounding.}
Video temporal grounding comprises two subtasks: moment retrieval and highlight detection.
Moment retrieval predicts the start and end timestamps of video moments that match a given query. 
Previous methods are typically categorized as either proposal-based~\cite{liu2021context,liu2020jointly,shao2018find,xiao2021boundary,xu2019multilevel,zhang2020learning,zhang2019cross} or proposal-free~\cite{chen2020rethinking,liu2022memory,liu2022reducing,lu2019debug,mun2020local,zeng2020dense} approaches.
Highlight detection, on the other hand, assigns saliency scores to video clips to identify relevant moments. 
While early works~\cite{gygli2016video2gif,xiong2019less,rochan2020adaptive,badamdorj2021joint} and benchmarks~\cite{song2015tvsum,sun2014ranking} do not consider textual queries, recent approaches incorporate text queries paired with the highlight moments, adopting several ranking-based strategies~\cite{liu2011query,liu2015multi,rochan2020adaptive,yu2018deep}.
Following the introduction of QVHighlights~\cite{lei2021detecting}, there has been growing interest in addressing these two tasks jointly. 
For instance, Moment-DETR~\cite{lei2021detecting} and subsequent works~\cite{moon2023query,jang2023knowing,sun2024tr,moon2023correlation} leverage the DETR~\cite{carion2020end} based architectures for temporal grounding.
Still, most existing methods rely on single-scale representations, limiting their ability to adapt to varying multi-scale granularity. 

\noindent\textbf{Multi-scale representation.}
Multi-scale representations have proven effective in modeling hierarchical semantics in both images~\cite{NIPS2002_c182f930,KolekWAKL23,HeidariKKAACM23} and videos~\cite{fan2021mvit,Karim0WS23,ZhangW00ZRML24}.
Feature Pyramid Networks (FPNs)~\cite{ghiasi2019fpn,lin2017feature,tan2020efficientdet,wang2021adaptive}, have proven effective in modeling hierarchical semantics for object detection.
In video understanding, prior works~\cite{WangLWY17,YangXSDZ20,YuCZFZ22} tried to adapt temporal FPN to capture both short- and long-range dependencies.
Similarly, in video temporal grounding, R$^2$-tuning~\cite{liu2024r} and FlashVTG~\cite{cao2025flashvtg} employ a temporal pyramid, but their scales are produced by uniform temporal downsampling, lacking semantic adaptivity.
Meanwhile, Superpixels~\cite{ren2003learning,achanta2012slic,van2012seeds} are widely adopted for grouping adjacent pixels into semantically coherent units, particularly in image segmentation.
We extend this concept to video temporal grounding by introducing SuperClips, grouped segments of video clips that capture both temporal adjacency and semantic proximity.


%% file: section/method.tex
\input{figures/main_fig}

\section{Method}

\subsection{Overall framework}
\label{sec:3.1}
Given the paired input $(\mathcal{V}, \mathcal{T})$, where the video $\mathcal{V}=\{v_i\}^{L_\text{vid}}_{i=1}$ is the sequence of $L_\text{vid}$ video clips denoted as $v_i$ and $\mathcal{T}= \{t_i\}^{L_\text{txt}}_{i=1}$ is a textual query of $L_\text{txt}$ tokens, the major objective of SCVTG is to identify the moments in $\mathcal{V}$ that match the given query.
In \emph{moment retrieval}, the model predicts a set of temporal boundaries, whereas \emph{highlight detection} requires the model to predict the clip-wise saliency score regarding $\mathcal{T}$.
Following the common practice, SCVTG addresses two tasks jointly through three modules: i) multi-modal fusion, ii) SuperClip pyramid, and iii) task-specific prediction heads. 
The overall architecture of SCVTG is illustrated in~\Cref{fig:main_fig}.

First, in the multi-modal fusion, we extract the modality-specific features with the pretrained vision and text encoders, followed by consecutive linear layers to match their dimensionality, resulting in $F^\text{vid} \in\mathbb{R}^{L_\text{vid} \times C}$ and $F^\text{txt} \in\mathbb{R}^{L_\text{txt} \times C}$. 
After that, we leverage \emph{Adaptive Cross-Attention} (ACA) from CG-DETR~\cite{moon2023correlation}, followed by \emph{self-attention} (SA) for generating video representations combined with textual information.
The resulting representation, denoted as $F\in \mathbb{R}^{L_\text{vid} \times C}$, serves as the input to the SuperClip pyramid.

\input{figures/method_fig}
\subsection{SuperClip pyramid}
\label{sec:3.2}
The standard temporal pyramid~\cite{liu2024r,cao2025flashvtg} is typically generated by uniform downsampling using CNNs.
While CNNs implicitly encode local context during downsampling, they lack an explicit mechanism for grouping semantically related clips or respecting unique events.
To address this limitation, the SuperClip pyramid constructs content-aware segments that capture both semantic coherence and temporal adjacency.
Note that since ground-truth moments are generally defined as semantically coherent and temporally continuous segments aligned with the query, such SuperClips provide effective initial candidates for precise temporal grounding as illustrated in~\Cref{fig:main_fig}.

Analogous to the concept of \emph{Superpixels}~\cite{ren2003learning} that group nearby pixels in an image into perceptually meaningful clusters, \emph{SuperClip} indicates the grouped segments of video clips based on their temporal adjacency and semantic proximity.
Formally, given input features $F^{l} =\{f^{l}_1,\dots,f^{l}_L\} \in \mathbb{R}^{L\times C}$ in $l$-th stage of the SuperClip pyramid, the representation of the $i$-th SuperClip can be defined as:
\begin{align}
   f^{l+1}_i =\mathcal{A}(\mathcal{C}^{l}_i), \quad \mathcal{C}^{l} = \mathcal{G}(F^{l}).
   \label{eq:scpyramid}
\end{align}
Here, $\mathcal{C}^{l} =\{\mathcal{C}^{l}_1, \dots, \mathcal{C}^{l}_{\lceil L/2 \rceil}\}$, where $\mathcal{C}^{l}_i \subset F^{l}$, denotes the set of disjoint SuperClips generated by the grouping operation $\mathcal{G}(\cdot)$, and $\mathcal{A}(\cdot)$ is the set aggregation function agnostic to the number of clips in the group.

\noindent\textbf{SuperClip grouping.} We start by defining the grouping operation $\mathcal{G}(\cdot)$.
One na\"ive strategy is to adopt the previous Superpixel algorithms for the temporal domain.
Yet, in practice, these classical Superpixel algorithms require either iterative refinement~\cite{achanta2012slic, van2012seeds} or a non-trivial graph search~\cite{shi2000normalized, felzenszwalb2004efficient}, leading to significant computational overhead (see supplement for runtime comparison). 

Instead, we efficiently generate SuperClips as follows.
Given the representation of video clips $F^{l} = \{f^l_1,\dots,f^l_{L}\}$ at each layer, we first select the anchor $F^{\alpha}$ and non-anchor $F^{\beta}$ clips alternately from the original sequence as: 
\begin{align}
    &F^{\alpha} = \bigl\{f^l_{2k-1}\ \bigl|\ k= 1,\dots,\bigl\lceil\tfrac{L}{2}\bigr\rceil \bigr\}
            \in \mathbb{R}^{\lceil \frac{L}{2}\rceil \times C},\\
    &F^{\beta}  =  \bigl\{f^l_{2k}\ \bigl|\ k = 1,\dots,\bigl\lfloor\tfrac{L}{2}\bigr\rfloor \bigr\}
  \in  \mathbb{R}^{\lfloor \frac{L}{2}\rfloor \times C}.
\end{align}
Then, to group the semantically similar consecutive clips, we assign a non-anchor clip $f^\beta_j \in F^{\beta}$ to a coherent adjacent anchor clip in $F^\alpha$ based on the cosine similarity as:
\begin{equation}
a_j =
\operatorname*{argmax}_{i\in\{j,\;j+1\}} \frac{f^\alpha_i \cdot f^\beta_j}{\|f^\alpha_i\| \|f^\beta_j\|},
\;\text{for} \; j = 1,\dots,\lfloor\tfrac{L}{2}\rfloor.
\end{equation}

Lastly, each anchor gathers all clips assigned to it, including itself, producing disjoint SuperClips:
\begin{equation}
\begin{aligned}
\mathcal{C}^l_i = \mathcal{G}(F^l)_i &= \{f^{\alpha}_{i} \} \cup \{f^{\beta}_{j}\mid a_j=i\}, \;  \\ &\text{for}\;  i = 1,\dots,\lfloor\tfrac{L}{2}\rfloor.
\end{aligned}
\end{equation}

Note that SuperClip grouping draws inspiration from classical SuperPixel algorithms, which consider both semantic similarity and spatial proximity to generate \emph{spatially contiguous groups}.
Consequently, SuperClip is designed to capture \emph{temporally continuous moments} with semantically coherent events.

\subsection{Temporally Adaptive Kernel for aggregation}
\label{sec:3.3}
In this section, we propose a novel aggregation process for SuperClips.
Although previous works~\cite{liu2024r,cao2025flashvtg} employ 1-D convolution for temporal downsampling, such an approach is not suitable for SuperClips.
As depicted in the upper part of \Cref{fig:method_fig}.(c), a fixed kernel cannot accommodate a variable number of constituent clips within a SuperClip.
Moreover, the convolutional kernel weights are agnostic to the varying durations and positions of the constituent clips, leading to suboptimal temporal aggregation.
Other alternatives, either defining multiple CNN kernels for different combinations of SuperClips or using permutation-invariant pooling, are also impractical due to excessive cost and the loss of temporal order, respectively.

Here, to achieve flexible aggregation adapting to arbitrary temporal durations and positions, we propose a \textit{\textbf{T}emporally \textbf{A}daptive \textbf{K}ernel} (\textbf{TA-Kernel}).
Inspired by Mixture of Experts (MoE)~\cite{JacobsJNH91}, TA-Kernel dynamically generates aggregation kernels based on temporal information.
Specifically, we generate a temporally adaptive kernel as a mixture of $B$ basis weights according to routing scores derived from temporal context as:
\begin{equation}
\hat{W}_i = \sum_{b=1}^{B} r_{ib} W_b, \quad 
\text{where} \quad 
r_i = \sigma\!\left(q_i K^\top\right),
\label{eq:PPA_1}
\end{equation}
where $r_i \in \mathbb{R}^{B}$ is the routing scores, the query $q_i = \phi(f^l_i)$ is the temporal encoding of the $i$-th clip, $K  \in \mathbb{R}^{B \times C}$ denotes a set of learnable expert keys of distinct experts, and $\sigma$ is the activation function. 
Consequently, an adaptive projection weight $\hat{W}_i$ is generated considering the temporal context of the clip.
For the positional embedding $\phi(\cdot)$, we concatenate the sinusoidal embeddings of the center and length of SuperClips, normalized by the total video length.
And, Softmax is used for activation $\sigma$, so $\hat{W}_i$ becomes a convex combination of weight bases as $\sum^B_{b=1} r_{ib} = 1$ (See~\Cref{sec:4.2} for more options for $\phi$, $\sigma$ ).
Then, the constituent clip feature is linearly projected using its dedicated weight $\hat{W}_i$, yielding temporally contextualized representations within the SuperClip as
\begin{equation}
\hat{f}_i = f^l_i \hat{W}_i.
\label{eq:PPA}
\end{equation}
Lastly, given the updated representations $\hat{f}_i$, we apply a pooling layer followed by an output projection as follows:
\begin{equation}
\begin{aligned}
   &\tilde{f}_j =  \mathcal{P}(\{\hat{f}_i | f^l_i \in \mathcal{C}^l_j\})\\
    &f^{l+1}_j = \text{Linear}(\tilde{f}_j) .
    \label{eq:PPA_2}
\end{aligned}
\end{equation}
Consequently, as illustrated in~\Cref{fig:main_fig}.(b), our aggregation can be viewed as generating a temporally adaptive kernel for each SuperClip to perform temporal-aware downsampling.
This process effectively summarizes SuperClips regardless of its group size, while considering the diverse temporal characteristics of its constituent clips.

\noindent\textbf{Low-rank approximation of weight basis.} 
The temporally adaptive kernel is an effective solution for the SuperClip aggregation.
We improve its efficiency by a simple low-rank approximation of weight bases.
Plain projection with TA-Kernel requires $B \times$ more learnable parameters for the expert weights $\{W_b\}_{b=1}^{B}$ compared to a standard linear layer. 
Further, generating clip-specific projection matrices $\hat{W} = \{\hat{W}_1, \dots, \hat{W}_L\} \in \mathbb{R}^{L \times C \times C}$, as defined in~\Cref{eq:PPA_1}, incurs a substantial computational cost of $\mathcal{O}(LBC^2)$ and memory cost of $\mathcal{O}(LC^2)$, where $L$, $B$, and $C$ denote the sequence length, the number of basis matrices, and the dimensionality, respectively.
Here, to mitigate memory and computational overhead, the basis matrix $W_b= U_b V_b$ is decomposed into two low-rank matrices $U_b \in \mathbb{R}^{C \times \rho}$ and $V_b \in \mathbb{R}^{\rho \times C}$, where $\rho \ll C$ is the low-rank.
To maintain the same number of parameters with a standard linear layer, we set $\rho = \frac{C}{2B}$.
We further reduce the computational complexity by reordering the multiplication, yielding an equivalent result, as follows:
\begin{equation}
\hat{f}_i = \sum^B_{b=1}r_{ib} ((f^l_iU_b)V_b).
\label{eq:PPA_3}
\end{equation}
Note that this requires $\mathcal{O}(LC^2)$ and $\mathcal{O}(LBC+C^2)$ computational and memory complexity, respectively.
The details for deriving the computational cost are provided in the supplement.

\subsection{Prediction heads}
\label{sec:3.4}
\noindent\textbf{Moment retrieval.}
Given the SuperClip pyramid $\{F^{1}, \dots, F^{K}\}$, where $F^{l}$ is the $l$-th level representation, we concatenate all levels and apply self-attention for multi-scale information exchange:
\begin{equation}
\tilde{F} = \operatorname{SA} \left( \operatorname{concat} \left( \{ F^l \}_{l=1}^{K} \right) \right).
\end{equation} 
Then, each level of $\tilde{F}$ is passed through convolutions for regression, and an additional MLP for classification as:
\begin{equation}
    c = \text{MLP}(\text{conv}(\tilde{F})), \quad m  = \text{conv}(\tilde{F}),
\end{equation}
where $c\in \mathbb{R}^{|\tilde{F}|}$,  $m\in \mathbb{R}^{|\tilde{F}| \times 2}$ denote the binary classification scores, \eg, 0 for foreground and 1 for background given text query, and the predicted moments, \eg, (start, end).

\noindent\textbf{Highlight detection.}
We set the saliency scores $S = \{s_1,\dots,s_{L_\text{vid}}\} \in\mathbb{R}^{L_\text{vid}}$ for input video clips as
\begin{equation}
    s_i = \frac{f^1_iW_1 \cdot gW_2}{\sqrt{C}}, \quad g = \sum_{l=1}^{K} w^l \left(\frac{1}{|F^l|} \sum_{i=1}^{|F^l|}f^l_i\right),
\end{equation}
where $g \in \mathbb{R}^{C}$ is the global representation weighted by $w\in\mathbb{R}^{K}$, and $W_1,W_2 \in \mathbb{R}^{C \times C}$.

\subsection{Relative saliency loss}
\label{sec:3.5}
Even among query-relevant clips, some serve as more distinctive highlights than others.
To encourage the model to better recognize such differences in importance, we here introduce a series of \emph{relative saliency losses}.

\noindent\textbf{Relative saliency margin loss.}
Unlike the previous margin loss~\cite{lei2021detecting}, which uses a fixed margin $\Delta$, we define the \textit{relative saliency margin} based on the difference in ground-truth saliency scores as:
\begin{equation}
\mathcal{L}_\text{\textbf{rs}-margin} = \max\left(0, \Delta(s^\text{gt}_\text{low}, s^\text{gt}_\text{high}) + s_\text{low} - s_\text{high} \right)
\end{equation}
where $s$ denotes the predicted saliency score, $s^\text{gt}$ is the corresponding ground-truth saliency score, and $\Delta(x,y) = x-y$ is relative saliency margin. We calculate the loss with a hard pair of the clips with the highest and lowest ground-truth saliency scores within the annotated moment, and an easy pair, consisting of one clip inside the ground-truth moment and another outside it.
\input{table/qvhighlight}

\noindent\textbf{Relative saliency rank-aware contrastive loss.} 
We further define the \textit{relative saliency scores} $\tilde{s}$ as
\begin{equation}
\tilde{s}_i = \frac{s^\text{gt}_i}{\sum_{j=1}^{L_{\text{vid}}}s^\text{gt}_j},
\end{equation}
which are obtained by normalizing the ground-truth saliency $s^\text{gt}$ over the entire video.
We introduce this relative saliency scores $\tilde{s}$ as guidance for the rank-aware contrastive loss~\cite{moon2023query}:
\begin{equation}
\begin{aligned}
    &\mathcal{L}_\text{\textbf{rs}-rank} =\\&-\frac{1}{|R|} \sum_{\mathbf{r} \in R} \sum^{L_\text{vid}}_{i=1} \mathbbm{1}_{\left[s^\text{gt}_i>\mathbf{r}\right]} \tilde{s}_i \log \left( \frac{\exp (s_i/ \tau)}{\sum^{L_\text{vid}}_{j=1} \exp(s_j /\tau)}\right),  
\end{aligned}
\end{equation}
where $\mathbbm{1}$ is the indicator, $R$ is the set of saliency levels, and $\tau=0.5$ is the temperature.

\noindent\textbf{Relative saliency NCE loss.} 
Similarly, we introduce relative saliency $\tilde{s}$ to NCE loss from R2-Tuning~\cite{liu2024r} as an auxiliary loss, defined as:
\begin{equation}
    \mathcal{L}_\text{\textbf{rs}-NCE} = -\sum^{L_\text{vid}}_{i=1} \tilde{s}_i \log \left( \frac{\exp (\text{sim}(f^1_i, g_\text{txt}) / \tau^\prime)}{\sum^{L_\text{vid}}_{j=1} \exp(\text{sim}(f_j, g_\text{txt}) / \tau^\prime)} \right),
\end{equation}
where $\text{sim}(f^1_i, g_\text{txt})$ is the cosine similarity between the $i$-th clip representation and $g_\text{txt}$, the global text embedding, and $\tau^\prime=0.07$.

Note that both the \emph{relative saliency rank-contrastive loss} and \emph{NCE loss} assign higher loss weights to clips having greater ground-truth saliency, thereby encouraging the model to align both the ordering of predicted saliency scores and cosine similarities with the ground-truth saliency ranking.

By integrating the relative saliency losses with focal loss~\cite{lin2017focal} for classification ($\mathcal{L}_{\text{cls}}$) and L1 loss for boundary regression ($\mathcal{L}_{\text{reg}}$), the overall training objective of SCVTG is 
\begin{equation}
\begin{aligned}
    \mathcal{L}_\text{total} =  &\lambda_\text{reg}\mathcal{L}_\text{reg} +\lambda_\text{cls}\mathcal{L}_\text{cls}+\lambda_\text{\textbf{rs}-margin}\mathcal{L}_\text{\textbf{rs}-margin}
\\
&+\lambda_\text{\textbf{rs}-rank}\mathcal{L}_\text{\textbf{rs}-rank}
+\lambda_\text{\textbf{rs}-NCE}\mathcal{L}_\text{\textbf{rs}-NCE},
\end{aligned}
\end{equation}
where each $\lambda$ is the coefficient for its corresponding loss.

%% file: figures/main_fig.tex
\begin{figure*}[t!]
\begin{center}
\includegraphics[width=0.95\textwidth]{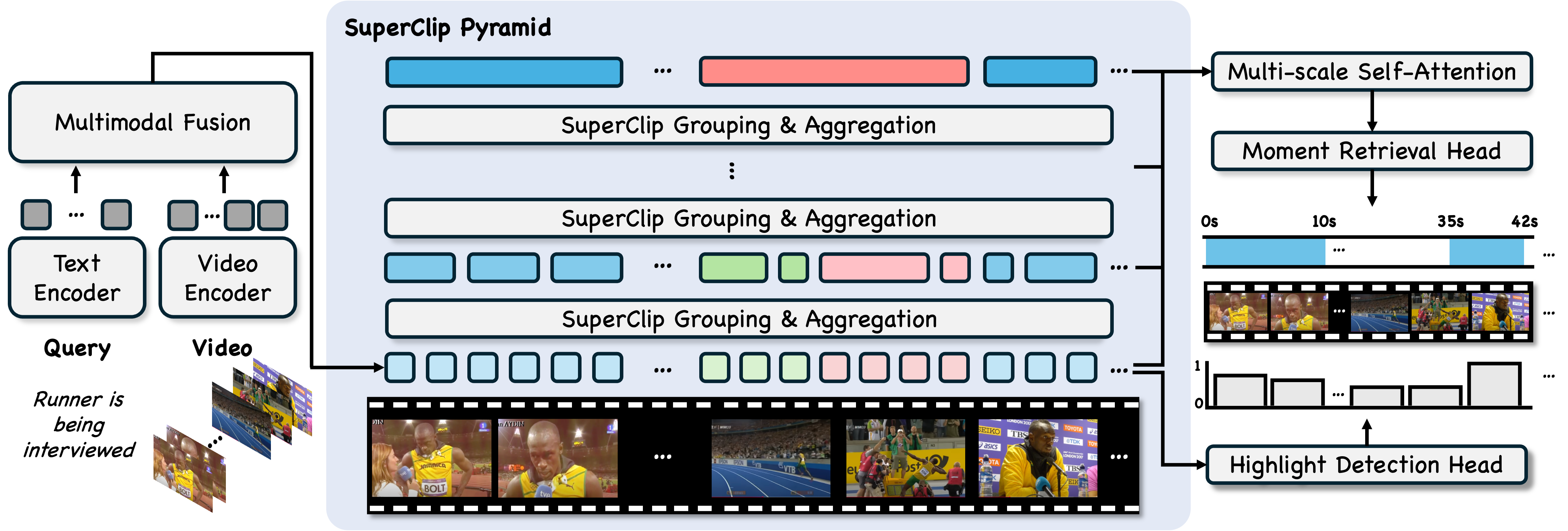}
\end{center}
\vspace{-15pt}
\caption{
\textbf{Overall architecture of SCVTG.}
After multi-modal fusion, video clip features are progressively grouped and aggregated to form a SuperClip pyramid with varying temporal granularity. 
Given this multi-level hierarchy capturing distinctive events, task-specific heads predict both foreground moments and clip-wise saliency scores.
}
\label{fig:main_fig}
\vspace{-10pt}
\end{figure*}

%% file: figures/method_fig.tex

\begin{figure*}[t!]
\begin{center}
\includegraphics[width=1.0\textwidth]{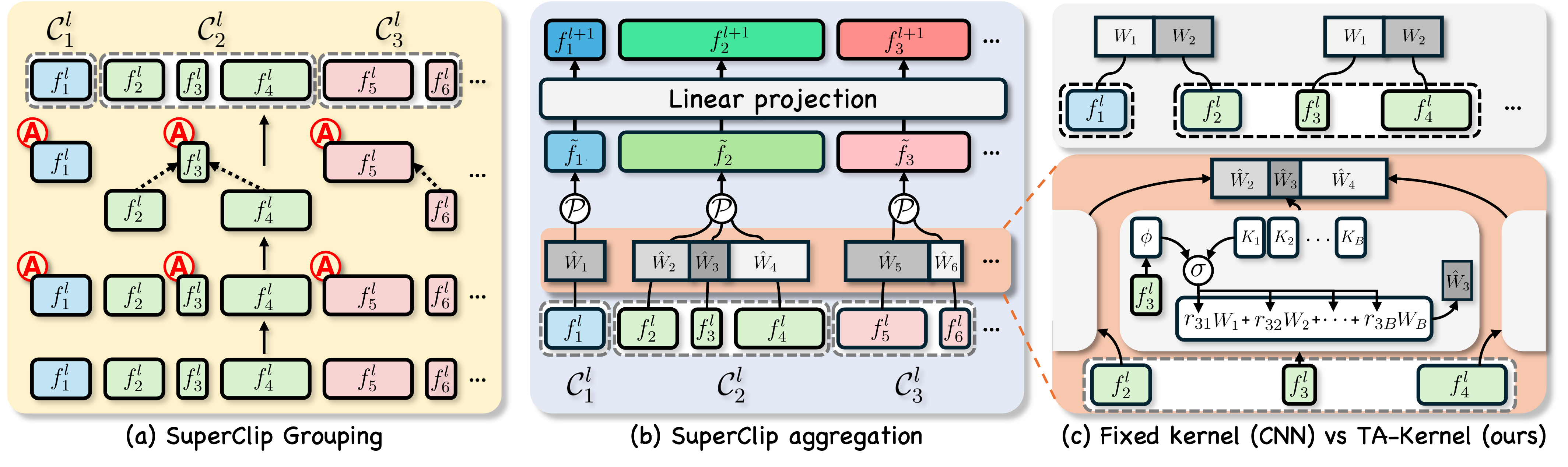}
\end{center}
\vspace{-20pt}
\caption{
\textbf{Illustration of SuperClip grouping and aggregation.}
(a) Given the input clips, a non-anchor clip is assigned to one of its adjacent anchor clips based on semantic similarity, forming coherent SuperClips.
(b) Next, each clip is projected using its corresponding weight generated by the temporally adaptive kernel (\textbf{TA-Kernel}), followed by pooling and a linear layer.
(c) Unlike fixed convolutional kernels, which cannot handle varying lengths or temporal characteristics, we generate TA-kernel weights for SuperClips, considering the temporal information ($\phi(f_i)$) of its constituent clips.
}
\label{fig:method_fig}
\vspace{-10pt}
\end{figure*}

%% file: table/qvhighlight.tex
\begin{table*}[t]
    \centering
    \small
    \setlength{\tabcolsep}{3pt}
    \begin{tabular}{l|ccccccc|ccccccc}
    \toprule
    \multicolumn{1}{c|}{\multirow{3}{*}[-3pt]{\textbf{Method}}} & \multicolumn{5}{c}{\textbf{MR Val}} & \multicolumn{2}{c|}{\textbf{HD Val}}  & \multicolumn{5}{c}{\textbf{MR Test}} & \multicolumn{2}{c}{\textbf{HD Test}}   \\
     & \multicolumn{2}{c}{R1} & \multicolumn{3}{c}{mAP} & Hit & mAP & 
     \multicolumn{2}{c}{R1} & \multicolumn{3}{c}{mAP} & Hit & mAP \\
     \cmidrule(lr){2-3} \cmidrule(lr){4-6} \cmidrule(lr){7-8}
     \cmidrule(lr){9-10} \cmidrule(lr){11-13} \cmidrule(lr){14-15}
      & @0.5 & @0.7 & @0.5 & @0.75 & Avg & @1 & Avg & @0.5 & @0.7 & @0.5 & @0.75 & Avg & @1 & Avg\\
    \midrule
    MomentDETR~\cite{lei2021detecting} & 53.94 & 34.84 & - & - & 32.20 & 55.55 & 35.65  & 52.89 & 33.02 & 54.82 & 29.40 & 30.73 & 55.60 & 35.69 \\
    UniVTG~\cite{lin2023univtg} & 59.74 & - & - & - & 36.13 & 61.81 & 38.83 & 58.86 & 40.86 & 57.60 & 35.59 & 35.47 & 60.96 & 38.20 \\
    UMT~\cite{liu2022umt} & 60.26 & 44.26 & 56.70 & 39.90 & 38.59 & 64.19 & 39.85 & 56.23 & 41.18 & 53.83 & 37.01 & 36.12 & 59.99 & 38.18\\
    MomentDiff~\cite{li2023momentdiff} &  - & - & - & - & - & - & -  & 58.21 & 41.48 & 54.57 & 37.21 & 36.84 & – & – \\
    MH-DETR~\cite{xu2024mh} & 60.84 & 44.90 & 60.76 & 39.64 & 39.26 & 61.74 & 38.77 & 60.05 & 42.48 & 60.75 & 38.13 & 38.38 & 60.51 & 38.22\\
    QD-DETR~\cite{moon2023query} & 62.68 & 46.66 & 62.23 & 41.82 & 41.22 &  63.03& 39.13  &62.40 & 44.98 & 62.52 & 39.88 & 39.86 & 62.40 & 38.94 \\
    EaTR~\cite{jang2023knowing} & 61.36 & 45.79 & 61.86 & 41.91 & 41.74 & 58.65  & 37.15& - & - & - &- & - & - & - \\
    CG-DETR~\cite{moon2023correlation} & 67.35 & 52.06 & 65.57 & 45.73 & 44.93 & 66.71  & 40.79 & 65.43 & 48.38 & 64.51 & 42.77 & 42.86 & 66.21  & 40.33 \\
    TR-DETR~\cite{sun2024tr} & 67.10& 51.48 & 66.27 & 46.42 & 45.09& 64.77 & 40.55 & 64.66 & 48.96 & 63.98 & 43.73 & 42.62 & 63.42 & 39.91 \\
    UVCOM~\cite{xiao2024bridging} &65.10 & 51.81 & - & - & 45.79 & 63.29 & 40.03& 63.55& 47.47& 63.37 & 42.67 & 43.18 & 64.20 & 39.74 \\
    R$^2$-Tuning~\cite{liu2024r} & 68.71 & 52.06 & - &-  &47.59 & 64.13 & 39.45 & 68.03 & 49.35 & 69.04 & 47.56 & 46.17& 64.20& 40.75  \\
    FlashVTG ~\cite{cao2025flashvtg} & 73.10 & 57.29 & 72.75 & 54.33 & 52.84 & 72.90 & 44.15  & 70.69 & 53.96 & 72.33 & \textbf{53.85} & 52.00 & 71.01 & 44.09 \\
    \midrule
    \rowcolor{cyan!20} \textbf{SCVTG} & \textbf{75.10} &\textbf{59.23} & \textbf{75.18} & \textbf{56.25} & \textbf{54.50} & \textbf{74.32} & \textbf{45.33} & \textbf{72.70} &\textbf{55.77} & \textbf{73.78} & 53.70 & \textbf{52.73} & \textbf{71.92} & \textbf{44.62} \\
    \bottomrule
    \end{tabular}
    \caption{\textbf{Video moment retrieval and highlight detection results on QVHighlights~\cite{lei2021detecting}.} }
    \label{tab:main} 
\end{table*}

%% file: section/exp.tex
\input{table/tvsum_youtubehl}

\section{Experiments}

In this section, we first validate the effectiveness of
SCVTG on video temporal grounding in~\Cref{sec:4.1}. Then,
we conduct comprehensive ablation studies in~\Cref{sec:4.2}.
Additional implementation details and experiments are provided in the supplementary materials.

\subsection{Video temporal grounding}
\label{sec:4.1}
\textbf{QVHighlights.}
\Cref{tab:main} presents the results on QVHighlights~\cite{lei2021detecting}, a prominent benchmark for video temporal grounding that jointly addresses moment retrieval and highlight detection.
Following FlashVTG~\cite{cao2025flashvtg}, we provide the results using InternVideo2-6B~\cite{wang2024internvideo2} as a backbone.
We show that SCVTG consistently outperforms all existing methods in most metrics, except mAP@0.75 on the test set.
Specifically, on the validation set, SCVTG brings substantial gains of 2.43\%, 1.92\%, 1.66\% at mAP@0.5, mAP@0.75, mAP@Avg in MR and 1.18\% mAP@Avg in HD over the second-best model, FlashVTG.
Similarly, in the test set, SCVTG achieves the best R1@0.5,0.7 of 72.7\%, 55.77\%, and mAP@Avg of 52.73.
We also observe that SCVTG brings substantial improvements of 0.91\% Hit@1 and 0.63\% mAP@Avg in highlight detection on the test set.
Overall, these results clearly demonstrate the effectiveness of SCVTG across both moment retrieval and highlight detection.

\input{figures/qual}
\noindent\textbf{TVSum and YouTube-HL.}
We compare the HD results on TVSum~\cite{song2015tvsum} and YouTube-HL~\cite{sun2014ranking} in \Cref{tab:tvsum_ythl_combined} using SlowFast~\cite{feichtenhofer2019slowfast} and CLIP~\cite{radford2021learning} as backbones following previous works.
The average mAP improvements over the second-best models are 1.3\% on TVSum and 1.4\% on YouTube-HL, demonstrating that our approach is also effective in highlight detection.
On TVSum, our SCVTG achieves state-of-the-art performance on 6 out of 10 domains and competitive results on the remaining domains.
SCVTG also maintains strong performance on YouTube-HL, achieving the highest average mAP on 4 out of 6 domains.

\input{table/charades_tacos}
\noindent\textbf{Charades-STA and TACoS.}
We also evaluate SCVTG on MR using SlowFast~\cite{feichtenhofer2019slowfast} and CLIP~\cite{radford2021learning} as backbones with two benchmarks: Charades-STA~\cite{GaoSYN17} and TACoS~\cite{RegneriRWTSP13}.
The results are summarized in~\Cref{tab:main2}.
Our method achieves the highest recall and mIoU on both datasets, surpassing the previous state-of-the-art models.
We demonstrate significant performance improvements in R1@0.7, improving 1.59\% on Charades-STA and 3.00\% on TACoS, respectively.
Including the results of R1@0.7 on QVHighlights, these experimental results validate the effectiveness of our SuperClip pyramid in capturing semantically aligned moments with higher precision.

\subsection{Qualitative results}
In~\Cref{fig:qual}, we provide qualitative results of our SuperClip Pyramid on QVHighlights~\cite{lei2021detecting}, which is constructed by progressively downsampling coherent clips.
Each box denotes clip features across pyramid levels, with red arrows marking the annotated ground truth.
The given video is segmented into 75 clips, with clip indices shown along the bottom. 
The SuperClip Pyramid captures distinctive events at multiple temporal granularities, effectively partitioning the video into semantically coherent segments across varying durations.
Especially, as shown in the figure, this hierarchical structure allows the model to form SuperClips that are perfectly aligned with the ground-truth moments, as illustrated in the figure.
By grounding queries on these semantically consistent and hierarchically organized SuperClips, our method achieves accurate temporal localization.
Additional qualitative results for ground-truth moments with varying durations are included in the supplementary material.


\subsection{Ablation studies}
\label{sec:4.2}
In~\Cref{tab:abl} and~\Cref{tab:abl2}, we ablate key components of SCVTG. 
Unless otherwise stated, we conduct all ablations on QVHighlights~\cite{lei2021detecting}.

\input{table/abl}

\input{table/abl2}

\noindent\textbf{Grouping method (a).}
We compare our SuperClip grouping with naive pairwise merging strategies that sequentially combine two adjacent tokens, and the similarity-based grouping method used in Token Merging (Bipartite Soft Matching).
As shown in the table, our SuperClip grouping demonstrates clear advantages over these approaches, highlighting the effectiveness of semantic coherence and temporal continuity for video temporal grounding.

\noindent\textbf{Temporal adaptive kernel (b).} 
We validate the advantage of TA-Kernel by comparing it with a simple 1D convolution used in~\cite{liu2024r,cao2025flashvtg} and aggregation with a simple linear instead of TA-Kernel.
We observe that the aggregation with TA-Kernel outperforms all baselines with a substantial margin.
We believe that SCVTG effectively generates adaptive kernel weights for downsampling the SuperClips.

\noindent\textbf{Input for routing function (c).}
Temporal information $\phi$ of the clip is crucial for generating the TA-Kernel.
We evaluate two variants of the sinusoidal positional embedding (PE): center and length (\textbf{c, l}), and start and end moments (\textbf{s, e}). 
For each variant, we experiment with two types of positional encoding: (1) the absolute position values normalized by the total video length, and (2) relative position values with respect to the corresponding SuperClip.
Overall, the absolute position values provide better results, with the encoding generated by center and length achieving the best performance. Moreover, it outperforms routing based on raw features, a prevalent option in MoE, with the gain of 1.31\%, 0.76\% mAP in MR and HD, respectively, highlighting the importance of temporal conditioning for SuperClip aggregation.

\noindent\textbf{Activation of routing function (d).} 
For an activation in the routing function $\sigma(\cdot)$, the table shows that \textbf{Softmax} consistently surpasses the no-activation (\textbf{None}) and \textbf{ReLU}, with a substantial mAP gap of +2.47\%, +3.06\%, respectively.

\noindent\textbf{Pooling function (e).}
After updating the clips with temporally parameterized weights, we pool the clips in each SuperClip.
Here, we compare four pooling strategies: simple min and max pooling (\textbf{Min}, \textbf{Max}), average pooling (\textbf{Avg}), and average pooling weighted by clip length (\textbf{W.Avg}). We observe that simple max pooling gives a consistent performance gain over other pooling functions.

\noindent\textbf{Relative saliency loss (f).} 
We further ablate the relative saliency losses, where the removed loss is replaced with its original one (\textit{e.g.}, fixed-margin loss or the rank-aware contrastive and NCE loss without relative saliency weighting).
Notably, removing any relative saliency loss term consistently leads to a performance drop. 
This validates that integrating relative saliency is effective for improving both moment retrieval and highlight detection.

\noindent\textbf{Low-rank approximation for weight basis (g).}
For comparison of efficiency, we report the latency within the SuperClip pyramid and peak memory usage with a batch size of 64 in \Cref{tab:abl2}.
As discussed in~\Cref{sec:3.3}, na\"ively computing clip-wise projection matrix $\hat{W}$ as in~\Cref{eq:PPA_1} incurs huge memory cost and relatively higher computational cost.
In contrast, our low-rank approximation described in \Cref{eq:PPA_3} successfully alleviates both latency and memory usage while maintaining competitive performance.

%% file: table/tvsum_youtubehl.tex
\begin{table*}[t]
    \centering
    \small
    \setlength{\tabcolsep}{3pt}
    \begin{tabular}{l |cccccccccc|c |cccccc|c}
    \toprule
    \multicolumn{1}{c|}{\multirow{2}{*}{\textbf{Method}}} & \multicolumn{11}{c|}{\textbf{TVSum}~\cite{song2015tvsum} } &  \multicolumn{7}{c}{\textbf{YouTube-HL}~\cite{sun2014ranking} } \\
     & \textbf{VT} & \textbf{VU} & \textbf{GA} & \textbf{MS} & \textbf{PK} & \textbf{PR} & \textbf{FM} & \textbf{BK} & \textbf{BT} & \textbf{DS} & \textbf{Avg.} & 
    \textbf{Dog} & \textbf{Gym.} & \textbf{Par.} & \textbf{Ska.} & \textbf{Ski.} & \textbf{Sur.} & \textbf{Avg.} \\
    \midrule
    LIM-S~\cite{xiong2019less} &55.9&42.9&61.2&54.0&60.4&47.5&43.2&66.3&69.1&62.6&56.3 
                 &57.9 &41.7 &67.0 &57.8 &48.6 &65.1 &56.4 \\
    SL-Module~\cite{xu2021cross} &86.5&68.7&74.9&86.2&79.0&63.2&58.9&72.6&78.9&64.0&73.3 
                      &70.8 &53.2 &77.2 &72.5 &66.1 &76.2 &69.3 \\
    Joint-VA~\cite{badamdorj2021joint} &83.7&57.3&78.5&86.1&80.1&69.2&70.0&73.0&\textbf{97.4}&67.5&76.3 
                     &64.5 &71.9 &80.8 &62.0 &73.2 &78.3 &71.8 \\
    UniVTG~\cite{lin2023univtg} &83.9&85.1&89.0&80.1&84.6&81.4&70.9&91.7&73.5&69.3&81.0 
                   &71.8 &76.5 &73.9 &73.3 &73.2 &82.2 &75.2 \\
    UMT~\cite{liu2022umt} &87.5&81.5&88.2&78.8&81.4&87.0&76.0&86.9&84.4&79.6&83.1 
                &65.9 &75.2 &\textbf{81.6} &71.8 &72.3 &82.7 &74.9 \\
    QD-DETR~\cite{moon2023query} & 88.2 & 87.4 & 85.6 & 85.0 & 85.8 & 86.9 & 76.4 & 91.3 & 89.2 & 73.7 & 85.0 & 72.2 & 77.4 & 71.0 & 72.7 & 72.8 & 80.6 & 74.4 \\
    CG-DETR~\cite{moon2023correlation} & 86.9 & 88.8 & 94.8 & 87.7 & 86.7 & 89.6 & 74.8 & 93.3 & 89.2 & 75.9 & 86.8 & 76.3 & 76.1 & 70.0 & 76.0 & 75.1 & 81.9 & 75.9 \\
    FlashVTG~\cite{cao2025flashvtg} &88.3&\textbf{94.3}&91.5&87.7&87.1&91.1&74.7&93.4&90.3&81.7&88.0 
                     & \textbf{76.5} & 76.1 & 69.4 & 74.1 & 73.1 & 83.0 &75.4 \\
    TR-DETR~\cite{sun2024tr} & 89.3 & 93.0 & 94.3 & 85.1 & 88.0 & 88.6 & \textbf{80.4} & 91.3 & 89.5 & 81.6 & 88.1 & - & - & -& -& -&- \\
    LLMEPET~\cite{jiang2024prior} &\textbf{90.8}&91.9&94.2&\textbf{88.7}&85.8&90.4&78.6&93.4&88.3&78.7&88.1 
                    &73.6 &74.2 &72.5 &75.3 &73.4 &82.5 &75.3 \\
    \midrule
    \rowcolor{cyan!20} \textbf{SCVTG} & \textbf{90.8} & 94.1 & \textbf{96.2} & 85.8 & \textbf{90.0} & \textbf{92.1} & 72.8 & \textbf{96.6} & 92.4 & \textbf{83.4} & \textbf{89.4} & 73.2 & \textbf{80.0} & 74.0 & \textbf{76.5} & \textbf{76.2} & \textbf{83.7} & \textbf{77.3} \\

    \bottomrule
    \end{tabular}

    \caption{\textbf{Highlight detection on TVSum~\cite{song2015tvsum} and YouTube-HL~\cite{sun2014ranking}}.}
    \label{tab:tvsum_ythl_combined} 
\end{table*}

%% file: figures/qual.tex
\begin{figure*}[t!]
\centering
\includegraphics[width=\textwidth]{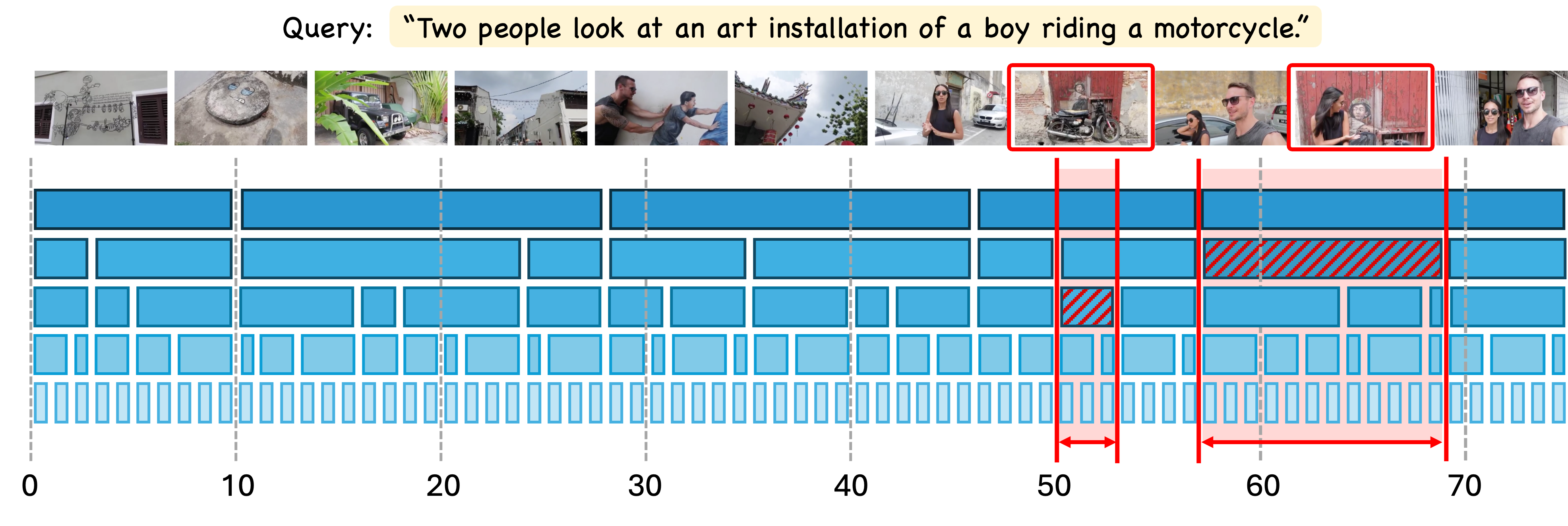}
\vspace{-5pt}
\caption{
\textbf{Qualitative results of SuperClip Pyramid.} The interval indicated by the red arrow denotes the ground-truth moments.
The SuperClip highlighted with a red hatched region perfectly aligns with the ground truth, serving as an ideal candidate for temporal grounding.
}
\label{fig:qual}
\end{figure*}

%% file: table/charades_tacos.tex
\begin{table}[t]
    \centering
    \small
    \begin{adjustbox}{width=\linewidth}
    \setlength{\tabcolsep}{1.5pt}
    \begin{tabular}{l|cccc |cccc}
    \toprule
    \multicolumn{1}{c|}{\multirow{2}{*}{\textbf{Method}}} & \multicolumn{4}{c|}{\textbf{Charades-STA}~\cite{GaoSYN17} } &  \multicolumn{4}{c}{\textbf{TACoS}~\cite{RegneriRWTSP13} } \\
    & @0.3 & @0.5 & @0.7 & mIoU & @0.3 & @0.5 & @0.7 & mIoU\\
    \midrule
    2D-TAN~\cite{zhang2020learning}  & 58.76 & 46.02 & 27.50 & 41.25 & 40.01 & 27.99 & 12.92  & 27.22 \\
    VSLNet~\cite{zhang2020span} & 60.30 & 42.69 & 24.14 & 41.58 & 35.54 & 23.54 & 13.15  & 24.99 \\
    MomentDETR~\cite{lei2021detecting}  & 65.83 & 52.07 & 30.59 & 45.54 & 37.97 & 24.67 & 11.97  & 25.49\\
    QD-DETR~\cite{moon2023query} & - & 57.31 & 32.55 & - & - & - & - & -\\
    TR-DETR~\cite{sun2024tr}& - & 57.61 & 33.52 & - & - &  - & - & -\\
    UniVTG~\cite{lin2023univtg} & 70.81 & 58.01 & 35.65 & 50.10 & 51.44 & 34.97 & 17.35 & 33.60 \\
    CG-DETR~\cite{moon2023correlation} & 70.43 & 58.44 & 36.34 & 50.25 & 52.32 & 39.59 & 22.16 & 36.55 \\
    LLMEPET~\cite{jiang2024prior}& 70.91 & - & 36.49 & 50.25 &52.73& - & 22.78 & 36.55\\
    R$^{2}$-Tuning~\cite{liu2024r} & 70.91 & 59.78 & 37.02 & 50.86 & 49.71 & 38.72 & 25.12 & 35.92 \\
    FlashVTG~\cite{cao2025flashvtg}  & - & 60.11 & 38.01 & - & 53.71 & 41.76 & 24.74 & 37.61\\
    \midrule
    \rowcolor{cyan!20} \textbf{SCVTG} & \textbf{71.67} & \textbf{60.67} & \textbf{39.60} & \textbf{51.63} & \textbf{56.11} & \textbf{43.64} & \textbf{27.74} & \textbf{39.58}\\
    \bottomrule
    \end{tabular}
    \end{adjustbox}
    \caption{\textbf{Moment retrieval on Charades-STA and TACoS.}}
    \label{tab:main2} 
\end{table}


%% file: table/abl.tex
\begin{table*}[t]
    \centering
    \small
    \begin{minipage}[t]{0.32\linewidth}
        \centering
        \setlength{\tabcolsep}{1pt}
        \begin{tabular}{l cccc}
        \toprule
        \multicolumn{1}{c}{\multirow{2}{*}[-3pt]{\textbf{Method}}} & \multicolumn{2}{c}{\textbf{MR}} & \multicolumn{2}{c}{\textbf{HD}}   \\
        \cmidrule(lr){2-3} \cmidrule(lr){4-5}
        & R1@0.7 & mAP & Hit@1 & mAP \\
        \midrule
        \multicolumn{5}{c}{\textbf{(a) Grouping method}}\\
        \midrule
        Pairwise & 58.58 & 53.83 & 74.00 & 44.71\\
        Similarity & 52.00 & 46.43 & 72.26 & 44.25 \\
        \rowcolor{cyan!20} SuperClip & \textbf{59.23} & \textbf{54.50} & \textbf{74.32} & \textbf{45.33} \\
        \\
        \midrule
        \multicolumn{5}{c}{\textbf{(b) Temporally adaptive kernel}}\\
        \midrule
        CNN &  57.52 & 53.33 & 73.74 & 44.40  \\
        Linear & 57.06 & 53.48 & 73.68 & 44.75  \\
        \rowcolor{cyan!20} TA-Kernel  & \textbf{59.23} & \textbf{54.50} & \textbf{74.32} & \textbf{45.33} \\\\
        \bottomrule
        \end{tabular}
    \end{minipage}%
    \hfill
    \begin{minipage}[t]{0.32\linewidth}
        \centering
        \setlength{\tabcolsep}{1pt}
        \begin{tabular}{l cccc}
        \toprule
        \multicolumn{1}{c}{\multirow{2}{*}[-3pt]{\textbf{Method}}} & \multicolumn{2}{c}{\textbf{MR}} & \multicolumn{2}{c}{\textbf{HD}}   \\
        \cmidrule(lr){2-3} \cmidrule(lr){4-5}
        & R1@0.7 & mAP & Hit@1 & mAP \\
        \midrule
        \multicolumn{5}{c}{\textbf{(c) Input for routing function}}\\
        \midrule
        Rel (\textbf{c, l}) & 57.68 & 52.01 & 73.29 & 44.46 \\
        Rel (\textbf{s, e}) & 57.29 & 51.51 & 73.30 & 44.51 \\
        \rowcolor{cyan!20} Abs (\textbf{c, l}) & \textbf{59.23} & \textbf{54.50} & \textbf{74.32} & \textbf{45.33} \\
        Abs (\textbf{s, e}) & 58.45 & 54.14 & 74.00 & 44.96  \\
        Feature & 57.81 & 53.19 & 73.23 & 44.57 \\
        \midrule
        \multicolumn{5}{c}{\textbf{(d) Activation of routing}}\\
        \midrule
        None & 57.94 & 52.03 & 73.61 & 44.46\\
        ReLU & 57.94 & 51.44 & 73.42 & 44.75 \\
        \rowcolor{cyan!20} Softmax& \textbf{59.23} & \textbf{54.50} & \textbf{74.32} & \textbf{45.33} \\

        \bottomrule
        \end{tabular}
    \end{minipage}%
    \hfill
    \begin{minipage}[t]{0.32\linewidth}
        \centering
        \setlength{\tabcolsep}{1pt}
        \begin{tabular}{l cccc}
        \toprule
        \multicolumn{1}{c}{\multirow{2}{*}[-3pt]{\textbf{Method}}} & \multicolumn{2}{c}{\textbf{MR}} & \multicolumn{2}{c}{\textbf{HD}}   \\
        \cmidrule(lr){2-3} \cmidrule(lr){4-5}
        & R1@0.7 & mAP & Hit@1 & mAP \\
        \midrule
        \multicolumn{5}{c}{\textbf{(e) Pooling function}}\\
        \midrule
        Avg & 58.39 & 52.96 & 73.48 & 44.30 \\
        W. Avg & 58.52 & 53.43 & 73.42 & 44.21\\
        Min & 57.68 & 51.89 & 73.48 & 44.53\\
        \rowcolor{cyan!20}Max & \textbf{59.23} & \textbf{54.50} & \textbf{74.32} & \textbf{45.33} \\
        \midrule
        \multicolumn{5}{c}{\textbf{(f) Relative saliency loss}}\\
        \midrule
        \rowcolor{cyan!20} w. all & \textbf{59.23} &          \textbf{54.50} & \textbf{74.32} & \textbf{45.33} \\
        w/o $\mathcal{L}_\text{\textbf{rs}-margin}$& 58.58 & 53.82 & 73.81 & 44.77 \\
        w/o $\mathcal{L}_\text{\textbf{rs}-rank}$  & 59.03 & 54.28 & 73.87 & 44.83 \\
        w/o $\mathcal{L}_\text{\textbf{rs}-NCE}$   & 58.90 & 54.07 & 73.74 & 44.61 \\

        \bottomrule
        \end{tabular}
    \end{minipage}%

    \caption{\textbf{Ablation studies on SCVTG.} The highlighted row indicates the default settings.}
    \label{tab:abl}
\end{table*}

%% file: table/abl2.tex
\begin{table}[t]
    \centering
    \small
    \centering
    \setlength{\tabcolsep}{1pt}
    \begin{tabular}{l cccc cc}
    \toprule
    \multicolumn{1}{c}{\multirow{2}{*}{\textbf{Method}}} & \multicolumn{2}{c}{\textbf{MR}} & \multicolumn{2}{c}{\textbf{HD }}  &\textbf{Latency} & \textbf{Memory}   \\
     & R1@0.7 & mAP & Hit@1 & mAP  & ms & M \\
    \midrule
    \multicolumn{7}{c}{\textbf{(g) Low-rank approximation for weight basis}}\\
    \midrule
    Original $W_b$  & \textbf{59.87} & 54.07 & \textbf{74.65} & 45.04 & 0.188 & 8222 \\
    \rowcolor{cyan!20} Low-rank $U_b,V_b$ & 59.23 & \textbf{54.50} & 74.32 & \textbf{45.33}  & 0.112 & 3978 \\
    \bottomrule
    \end{tabular}
    \caption{\textbf{Ablation studies on low-rank approximation.}}
    \label{tab:abl2} 
\end{table}

%% file: section/conclusion.tex
\section{Conclusion}
In this paper, we propose SCVTG, a unified framework for video temporal grounding that jointly addresses moment retrieval and highlight detection.
To better ground real-world events with varying durations, we proposed SuperClips, semantically coherent segments generated by grouping semantically similar and adjacent video clips.
Additionally, with the introduction of a TA-kernel into the aggregation, SCVTG effectively aggregates variable-length SuperClips by assigning temporally conditioned projection matrices to each constituent clip, thereby embedding sequential relationships into the resulting representation.
We also proposed a series of novel relative saliency losses to better reflect the importance differences among clips.
SCVTG achieves state-of-the-art performance across all benchmarks presented in this paper, validating the effectiveness of hierarchical grouping of semantically coherent segments for fine-grained video temporal grounding.

%% file: section/appendix.tex

\renewcommand{\thetable}{\Alph{table}}
\renewcommand{\thefigure}{\Alph{figure}}

\input{table/implementations_sup}

\section*{Appendix}
\section{Experimental Settings}
\textbf{Datasets.}
\textbf{QVHighlights}~\cite{lei2021detecting} consists of 10,310 YouTube videos, each paired with a manually annotated text query. 
It is designed for joint moment retrieval and highlight detection, where test set evaluations are available at the test server on the Codalab competition platform.
\textbf{Charades-STA}~\cite{GaoSYN17} is a moment retrieval dataset containing 9,848 videos with daily indoor activities and 16,128 human-written query texts. 
\textbf{TACoS}~\cite{RegneriRWTSP13} includes 127 cooking videos from the MPII Cooking Composite Activities dataset~\cite{rohrbach2012script}, with 18,818 queries and the corresponding moments are annotated.
\textbf{TVSum}~\cite{song2015tvsum} is a highlight detection dataset comprising 10 video categories, each containing 5 videos.
\textbf{YouTube-HL}~\cite{sun2014ranking} consists of 433 videos from six activity domains, \textit{e.g.}, dog, gymnastics, parkour, skating, skiing, and surfing, where the domain names are used as text queries for highlight detection.

\noindent\textbf{Evaluation metrics.}
For moment retrieval on QVHighlights, we report Recall@1 at IoU thresholds of 0.5 and 0.7, as well as mean Average Precision (mAP) at IoU thresholds of 0.5 and 0.75.
Following \cite{lei2021detecting,sun2024tr}, we also report average mAP by averaging mAPs of 10 uniformly sampled IoU thresholds from 0.5 to 0.95.
To evaluate highlight detection on QVHighlights, we use mAP and HIT@1.
For moment retrieval on Charades-STA and TACoS, we report mIoU and Recall@1 at IoU thresholds of 0.3, 0.5, and 0.7.
For highlight detection on TVSum and YouTube-HL, we use top-5 mAP as the primary evaluation metric.
%

\noindent\textbf{Implementation details.}
The architectural configurations and detailed hyperparameter settings of SCVTG are summarized in~\Cref{tab:imp} and~\Cref{tab:imp2}, respectively.
The architectural design is kept consistent across all datasets, except for the pyramid depth. The pyramid depth is adjusted considering the average video length in each dataset. Specifically, longer videos require a deeper pyramid to capture extended events, while shorter videos do not benefit from excessive depth.
Also, following the common practice in video temporal grounding, we perform grid search only on the commonly tuned parameters, such as epochs, batch size, learning rate, weight decay, and loss coefficients.
We report the latency and peak memory usage in ablation studies, with all evaluations conducted on a single GeForce RTX 3090 with a batch size of 64.
Under this setup, it only takes about 2 hours to train SCVTG on QVHighlights for 200 epochs.

\section{More Experimental Results}


\input{table/various_length}
\subsection{Video temporal grounding with various moment lengths}
We compare the performance of SCVTG with previous methods across different ground-truth moment lengths on QVHighlights~\cite{lei2021detecting}.
Specifically, we consider four cases:
(i) short (0–10 seconds),
(ii) medium (10–30 seconds),
(iii) long (30–150 seconds), and
(iv) all (0–150 seconds).
The results are summarized in~\Cref{tab:var}.
Overall, SCVTG consistently outperforms prior methods regardless of the length of the ground-truth moments, highlighting the effectiveness of the SuperClip pyramid in generating semantically coherent segments having various temporal granularities.

\subsection{Sensitivity analyses on temporally adaptive kernel}
To understand the sensitivity of the temporally adaptive kernel to the number of weight basis $B$ and the dimension of low rank $r$, we analyze SCVTG with varying parameters.
As discussed in the main paper, given the hidden dimension $C=256$, we set the number of weight bases $B$ and the low-rank dimension $r$ to be $2Br = 256$, matching the parameter count of a standard linear layer.
Specifically, $B=4$ and $r=32$ are used for default values.
In the sensitivity analysis, we vary one parameter while keeping the other fixed at its default.
Empirically, we observe that SCVTG yields satisfactory performance with the default values, and increasing either $B$ or $r$ beyond this point offers no significant performance improvement despite the inevitably increasing parameter count and computational cost.
\input{table/sensitivity_sup}
\input{table/superpixels}
\subsection {Runtime comparison with Superpixel algorithm}
As discussed in the main paper, conventional Superpixel algorithms incur substantial computational overhead. Here, we compare the runtime of our grouping method with two representative SuperPixel algorithms, SLIC~\cite{achanta2012slic} and SEEDS~\cite{van2012seeds}, for constructing the SuperClip pyramid.
As shown~\Cref{tab:superpixel}, SLIC exhibits about 3× higher latency and SEEDS incurs nearly 27× greater computational cost compared to SuperClip grouping.
Compared to these baselines, our SuperClip grouping achieves significantly lower latency. Moreover, although SLIC is relatively faster than SEEDS, it often fails to preserve temporal connectivity within SuperClips, leading to orphan segments as discussed in~\cite{achanta2012slic}.
This is particularly undesirable, as temporal adjacency is critical for accurate moment retrieval.
Overall, SCVTG performs the fastest grouping while consistently preserving temporal continuity within SuperClips.

\section{Analysis of Weight Basis Selection}
To verify whether the model utilizes the expert weight bases selectively based on the temporal information of each clip, we analyze the routing scores using the model trained on QVHighlights~\cite{lei2021detecting}.
In our framework, the input for calculating routing scores is conditioned on both the center position and the temporal length of each clip.
To disentangle the effects of these two factors, we separately vary each component while keeping the other fixed.
Specifically, we set the length to $1/L$ (\textit{i.e.}, minimum length normalized by the total video length) when analyzing the center position, and fix the center to 0.5 (\textit{i.e.}, the middle of the video) when analyzing the effect of clip length.
\Cref{fig:routing1,fig:routing2} visualizes how routing scores change.

Interestingly, we observe clear transitions in the dominant expert weights across different positions and lengths, indicating that the model leverages the positional encoding to route clips to distinct expert bases.
These transitions change smoothly as the center position or clip length varies, indicating that certain weight bases are more likely to be selected depending on the position and length.
This selective and smooth activation indicates that each part of the video benefits from a different projection, and the model learns to leverage the expert weight bases selectively based on temporal position.

\input{figures/selection_sup}

\input{figures/supp-qual}

\section{Complexity Comparison.}

In the default temporally adaptive kernel, we first generate a clip-wise projection matrix $\hat{W} \in \mathbb{R}^{L\times C \times C}$, and apply it to its corresponding clip.
Given the $B$ expert weight basis $W\in\mathbb{R}^{B \times C \times C}$, the clip-wise routing weights $r \in \mathbb{R}^{L \times B}$, and the representation of the clips $F^l= \{f^l_i\}^L_{i=1} \in \mathbb{R}^{L \times C}$, we can write this process in the tensor contraction notation (\textit{i.e.}, einstein summation notation) as follows:
\begin{align}
    \hat{W} &= \text{contract}(LB, BCD \rightarrow LCD)(r, W), \label{eq:sup1} \\
    \hat{F} &= \text{contract}(LC, LCD \rightarrow LD)(F^l, \hat{W}), \label{eq:sup2}
\end{align}
where we denote $D=C$ to emphasize that the second contraction corresponds to standard matrix multiplication between each clip vector and its corresponding weight matrix.
The computational complexity of \Cref{eq:sup1} is \( \mathcal{O}(LBC^2) \), as each of the \( L \) clips performs a weighted sum over \( B \) basis matrices of size \( C \times C \). The cost of \Cref{eq:sup2} is \( \mathcal{O}(LC^2) \), corresponding to \( L \) independent matrix multiplications between \( C \)-dimensional vectors and \( C \times C \) matrices.  
Therefore, the total computational cost is $\mathcal{O}(LBC^2 + LC^2) = \mathcal{O}(LBC^2)$.
For memory complexity, the input and intermediate tensors require $\mathcal{O}(BC^2)$, $\mathcal{O}(LB)$, $\mathcal{O}(LC)$, $\mathcal{O}(LC^2)$, $\mathcal{O}(LC)$ for $W, r, F^{l}, \hat{W}, \hat{F}$, respectively.
Here, $\hat{W}$ dominates the total memory usage considering $B \ll L,C$.
Hence, the overall memory complexity is $\mathcal{O}(LC^2)$.

In contrast, the low-rank approximation of weight basis using $U,V \in \mathbb{R}^{B\times C \times \rho }, \mathbb{R}^{B \times \rho  \times C}$ is computed as
\begin{align}
    \bar{F} &= \text{contract}(LC, BC\rho  \rightarrow LB\rho )(F^{l}, U), \label{eq:sup3} \\
    \tilde{F} &= \text{contract}(LB\rho , B\rho C \rightarrow LBC)(\bar{F} , V), \label{eq:sup4} \\
    \hat{F}^{l} &= \text{contract}(LB, LBC \rightarrow LC)(r , \tilde{F}). \label{eq:sup5}
\end{align}
In this approach, each input clip is first projected through all low-rank basis components using $U,V$ as \Cref{eq:sup3,eq:sup4}, where each step incurs a computational cost of $\mathcal{O}(LBC\rho )$. Then, the final weighted aggregation in \Cref{eq:sup5} requires $\mathcal{O}(LBC)$.
Therefore, the total computational cost is $\mathcal{O}(2LBC\rho  + LBC)$.
Assuming the low-rank dimension $r$ is chosen such that $2B\rho  = C$ to match the parameter count of a standard linear layer, the total cost simplifies to $\mathcal{O}(LC^2 + LBC)$, which becomes $\mathcal{O}(LC^2)$ given $B \ll C$.
Also, the memory cost to store intermediate steps in~\Cref{eq:sup3,eq:sup4,eq:sup5} is $\mathcal{O}(LB\rho  + LBC +  LC) = \mathcal{O}(LBC)$, with $\rho  \ll C$.
Along with the inputs $\mathbf{w} \in\mathbb{R}^{L\times B}, U \in \mathbb{R}^{B \times C \times \rho },V \in \mathbb{R}^{B \times \rho  \times C},F^{(l)} \in \mathbb{R}^{L \times C}$, which incurs $\mathcal{O}(LB+ 2BC\rho  + LC) = \mathcal{O}(2BC\rho +LC) = \mathcal{O}(C^2+LC)$, the total space complexity becomes $\mathcal{O}(LBC+C^2+LC) = \mathcal{O}(LBC + C^2)$.
Compared to the standard implementation in \Cref{eq:sup1,eq:sup2}, the low-rank approximation variant in \Cref{eq:sup3,eq:sup5} significantly reduces the computational and memory costs. 
The low-rank version reduces the dominant computational cost from \( \mathcal{O}(LBC^2) \) to \( \mathcal{O}(LC^2) \). 
Furthermore, the intermediate memory usage is also reduced from \( \mathcal{O}(LC^2) \) to $\mathcal{O}(LBC + C^2)$.

\section{Qualitative Results of SuperClip Pyramid}
In this section, we provide qualitative results of our SuperClip Pyramid on QVHighlights~\cite{lei2021detecting}, illustrating how our SCVTG capture diverse temporal structures.
As visualized in~\Cref{fig:supp-qual}, each box represents a clip feature across pyramid levels, while the red arrows denote the annotated ground-truth moments. 
Each video is uniformly divided into 75 clips.
Through progressive grouping of adjacent clips into SuperClips, our method adaptively constructs semantically coherent segments that vary in length, ranging from short actions to long-term activities, without relying on fixed temporal intervals.
Overall, these visualizations demonstrate that our SuperClips form a content-aligned hierarchy capable of representing both fine-grained and long-range dynamics, resulting in precise and context-aware moment localization.
\clearpage

%% file: table/implementations_sup.tex
\twocolumn[
\begin{center}
    \centering
    \small
    \setlength{\tabcolsep}{3pt}
    \begin{tabular}{l |cccc|ccc|ccc}
    \toprule
    \multicolumn{1}{c|}{\multirow{2}{*}{\textbf{Dataset}}} & \multicolumn{4}{c|}{\textbf{Multi-modal fusion}}  & \multicolumn{3}{c|}{\textbf{SuperClip pyramid }} & \multicolumn{3}{c}{\textbf{Prediction heads}}\\
     & D-Enc & ACA & SA & $L_{\text{dummy}}$& $K$ & $B$ & $r$ & Conv (cls) & MLP (cls) & Conv (reg) \\
    \midrule
    QVHighlights~\cite{lei2021detecting} & 2 & 6 & 3 & 40 & 5 & 4 & 32 & 1 & 5 & 2 \\
    Charades-STA~\cite{GaoSYN17} & 2 & 6 & 3 & 40 & 5 & 4 & 32 & 1 & 5 & 2 \\
    TACoS~\cite{RegneriRWTSP13} & 2 & 6 & 3 & 40 & 2 & 4 & 32 & 1 & 5 & 2 \\
    TVSum~\cite{song2015tvsum} & 2 & 6 & 3 & 40 & 3 & 4 & 32 & 1 & 5 & 2 \\
    YouTube-HL~\cite{sun2014ranking} & 2 & 6 & 3 & 40 & 4 & 4 & 32 & 1 & 5 & 2 \\
    \bottomrule
    \end{tabular}
    \captionof{table}{\textbf{Architectural configuration of SCVTG.} D-Enc: the number of dummy encoder layers for ACA~\cite{moon2023correlation}. ACA: the number of adaptive cross-attention layers. SA: the number of self-attention layers. $L_\text{dummy}$: the number of dummy tokens. $K$: the number of levels in the SuperClip pyramid. $B$: the number of expert weights. $r$: the low rank dimension. Conv (cls), MLP(cls): the number of convolutional, MLP layers for box classification. Conv (reg): the number of convolutional layers for box regression.}
    \label{tab:imp} 
\end{center}

\begin{center}
    \centering
    \small
    \setlength{\tabcolsep}{4pt}
    \begin{tabular}{c |ccccc}
    \toprule
    Hyperparameters
     & QVHighlights & Charades-STA & TACoS & TVSum & YouTube-HL \\ 
    \midrule
    Epochs & 200 & 100 & 100 & 400 & 1000 \\
    Batch size & 64 & 128 & 32 & 4 & 4 \\
    Learning rate & 5e-4 & 1.5e-4 & 2e-4 & 3e-3 & 3e-4 \\
    Weight decay & 1e-4 & 1e-4 & 1e-4 & 1e-1 & 1e-4 \\
    Hidden dimension $C$ & 256 & 256 & 256 & 256 & 256 \\
    Gradient clipping & 0.1 & 0.1 & 0.1 & 0.1 & 0.1 \\
    Optimizer &  AdamW &  AdamW &  AdamW &  AdamW &  AdamW \\
    Optimizer momentum & (0.9, 0.999) & (0.9, 0.999) & (0.9, 0.999) & (0.9, 0.999) & (0.9, 0.999)\\ 
    $\lambda_\text{reg}$ & 1 & 1 & 1 & 0.2 & 0.2\\
    $\lambda_\text{cls}$ & 5 & 5 & 5 & 10 & 5 \\
    $\lambda_\text{margin}$ & 0.8 & 0.8 & 0.8 & 1 & 0.1 \\
    $\lambda_\text{rank}$ & 0.8 & 0.8 & 0.8  & 1 & 0.1 \\
    $\lambda_\text{NCE}$ & 0.1 & 0.01 & 0.05 & 0.45 & 0.2 \\
    EMA decay rate & 0.998 & 0.998 & 0.998 & 0.998 & 0.998 \\
    NMS IoU threshold & 0.7  & 0.7  & 0.7  & 0.7  & 0.7 \\
    \bottomrule
    \end{tabular}
    \captionof{table}{\textbf{Implementation details of SCVTG hyperparameters.}}
    \label{tab:imp2} 
\end{center}
]

%% file: table/various_length.tex
\begin{table}[t!]
    \centering
    \small
    \setlength{\tabcolsep}{6pt}
    \begin{tabular}{l |cccc}
    \toprule
    \multicolumn{1}{c|}{\textbf{Method}} & \textbf{Short} & \textbf{Middle} & \textbf{Long} & \textbf{All}\\
    \midrule
    UMT~\cite{liu2022umt} & 5.02 &42.22 & 42.84 & 38.59 \\
    UniVTG~\cite{lin2023univtg} & 8.64 &39.90 & 37.82 & 36.13 \\
    CG-DETR~\cite{moon2023correlation} & 10.58 &48.30 & 49.36 & 44.93\\
    LLMEPET~\cite{jiang2024prior} & 10.58 &48.30 & 49.36 & 44.93\\
    $R^2$-Tuning~\cite{liu2024r} & 12.62 & 48.56 & 55.42 & 47.86 \\
    FlashVTG~\cite{cao2025flashvtg} & 15.65 & 56.54 & 56.19 & 52.84\\
    \midrule
    SCVTG & \textbf{17.63} & \textbf{57.05} & \textbf{57.60} & \textbf{54.50} \\
    \bottomrule
    \end{tabular}
    \caption{\textbf{Results according to the various moment lengths.}  Results are reported using mAP@Avg of moment retrieval.}
    \label{tab:var} 
\end{table}

%% file: table/sensitivity_sup.tex
\begin{table}[t!]
    \centering
    \small
    \setlength{\tabcolsep}{5pt}
    \centering
    \begin{tabular}{c|ccccc}
    \toprule
    \# $B$ & 2 & \cellcolor{cyan!20} 4 & 8 & 16 & 32\\
    \midrule
    mAP & 54.14 & \cellcolor{cyan!20}54.50 & 54.48 & 54.57 & 54.24 \\
    \bottomrule
    \end{tabular}
    \centering
    \begin{tabular}{c |ccccc}
    \toprule
    \# $r$ & 8 & 16 & \cellcolor{cyan!20}32 & 64 & 128\\
    \midrule
    mAP & 52.81 & 53.81 & \cellcolor{cyan!20} 54.50 & 54.52 & 54.35\\
    \bottomrule
    \end{tabular}
    \caption{\textbf{Sensitivity analysis on the number of weights basis $B$ and the dimension of low rank $r$ in positional parameterization.}} 
    \label{tab:sen} 
\end{table}

%% file: table/superpixels.tex
\begin{table}[t]
    \centering
    \small
    \centering
    \setlength{\tabcolsep}{4pt}
    \begin{tabular}{l ccc}
    \toprule
    Method & SLIC~\cite{achanta2012slic} & SEEDS~\cite{van2012seeds} & \cellcolor{cyan!20} SuperClip \\
    \midrule
    Latency(ms) & 0.355 & 3.069 & \cellcolor{cyan!20}  \textbf{0.112} \\
    \bottomrule
    \end{tabular}
    \caption{\textbf{Runtime comparison of grouping method.}}
    \label{tab:superpixel} 
\end{table}

%% file: figures/selection_sup.tex
\begin{figure}[t!]
  \centering

  \begin{minipage}[b]{0.48\linewidth}
    \centering
    \includegraphics[width=\linewidth]{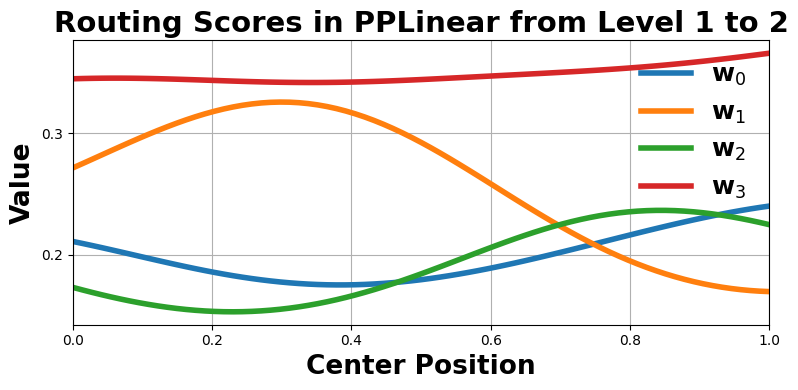}
  \end{minipage}
  \hfill
  \begin{minipage}[b]{0.48\linewidth}
    \centering
    \includegraphics[width=\linewidth]{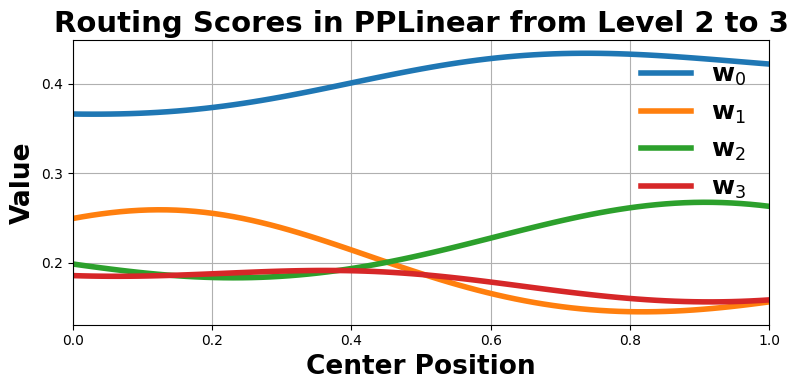}
  \end{minipage}

  \begin{minipage}[b]{0.48\linewidth}
    \centering
    \includegraphics[width=\linewidth]{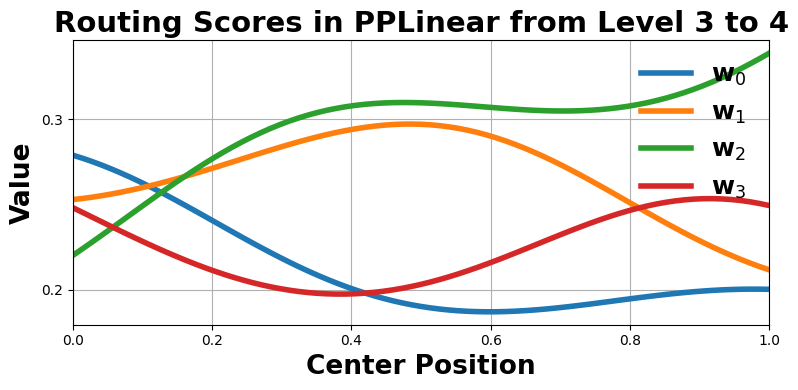}
  \end{minipage}
  \hfill
  \begin{minipage}[b]{0.48\linewidth}
    \centering
    \includegraphics[width=\linewidth]{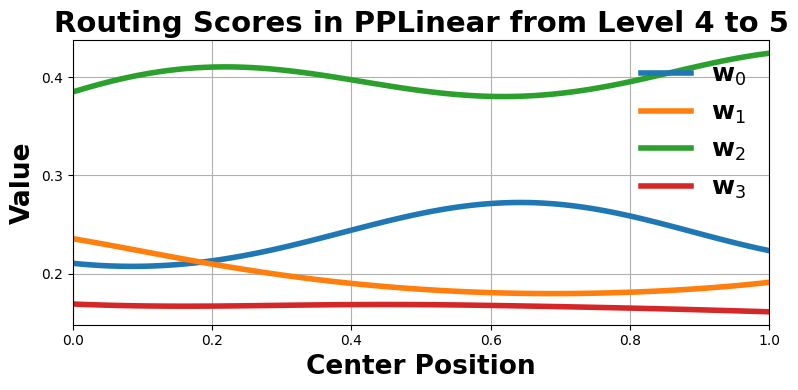}
  \end{minipage}

  \caption{Routing scores from different center positions  across pyramid levels.}
  
  \label{fig:routing1}

  \centering

  \begin{minipage}[b]{0.48\linewidth}
    \centering
    \includegraphics[width=\linewidth]{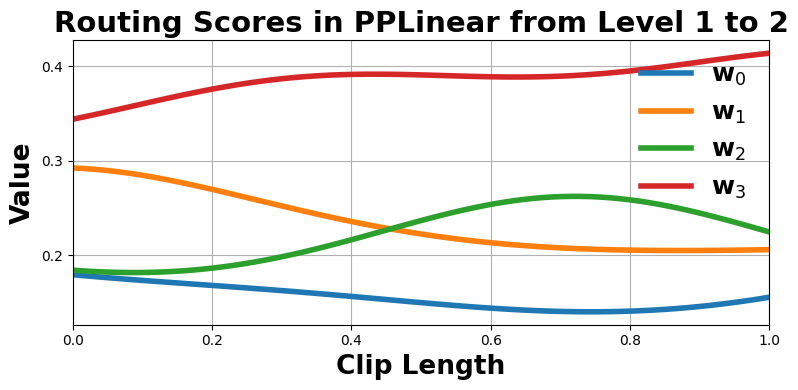}
  \end{minipage}
  \hfill
  \begin{minipage}[b]{0.48\linewidth}
    \centering
    \includegraphics[width=\linewidth]{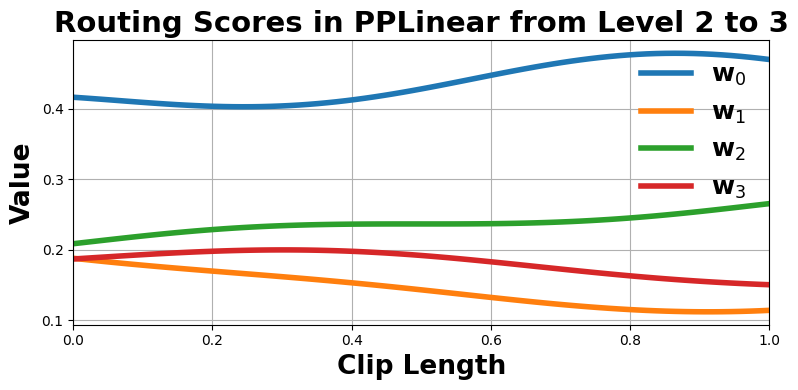}
  \end{minipage}

  \begin{minipage}[b]{0.48\linewidth}
    \centering
    \includegraphics[width=\linewidth]{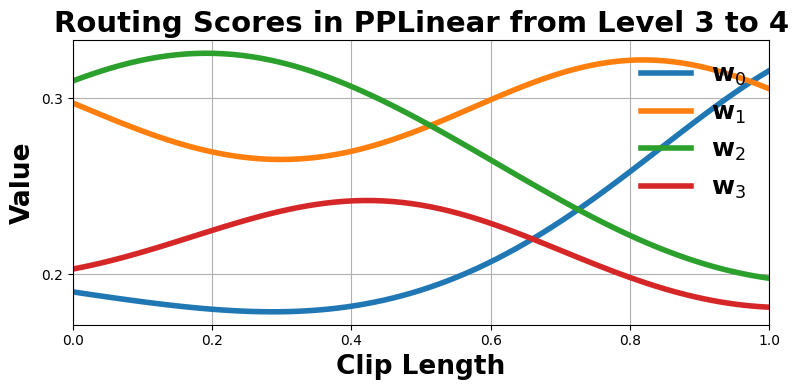}
  \end{minipage}
  \hfill
  \begin{minipage}[b]{0.48\linewidth}
    \centering
    \includegraphics[width=\linewidth]{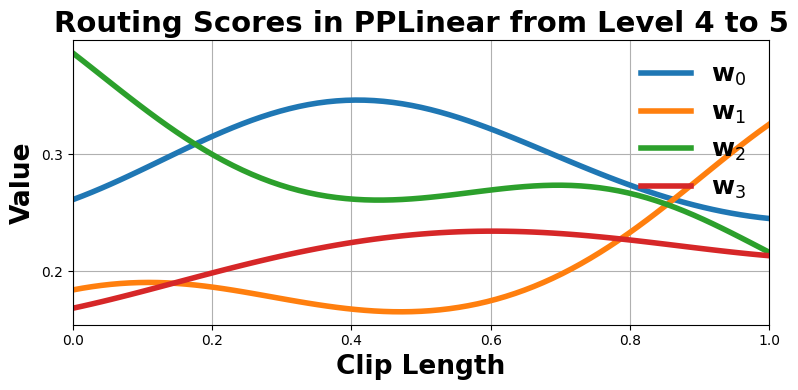}
  \end{minipage}

  \caption{Routing scores from different clip lengths across pyramid levels.}
  
  \label{fig:routing2}
  \vspace{-15pt}
\end{figure}

%% file: figures/supp-qual.tex
\begin{figure*}[t!]
\centering
\includegraphics[width=\textwidth]{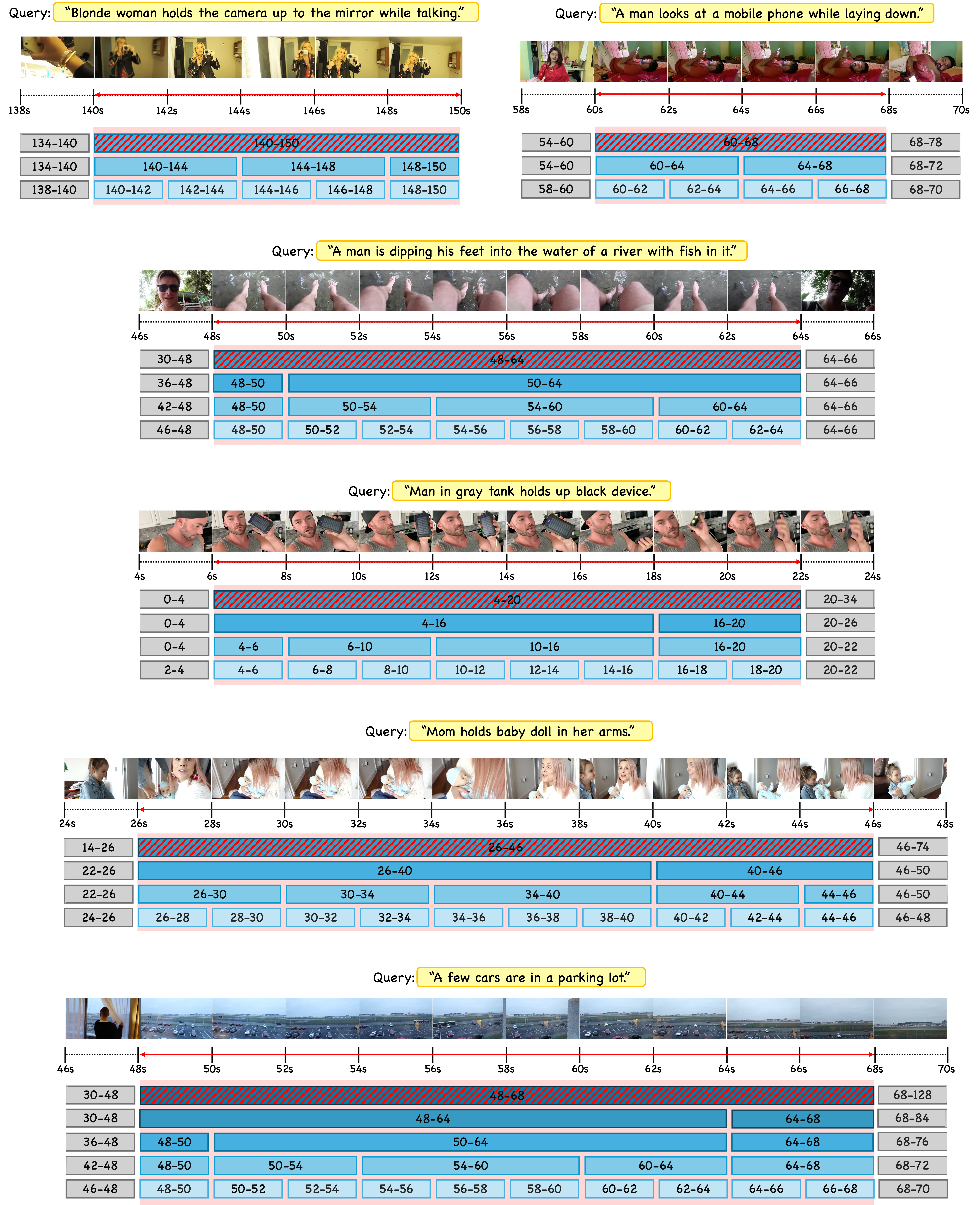}
\vspace{-5pt}
\caption{
\textbf{Qualitative results of SuperClip Pyramid.} The red arrows indicate the ground truth moments.
}
\label{fig:supp-qual}
\end{figure*}